\documentclass[letterpaper, 10 pt, conference]{ieeeconf-modified}

\IEEEoverridecommandlockouts

\usepackage[numbers]{natbib}
\usepackage{graphicx}
\graphicspath{ {./figures} }
\usepackage{siunitx}
\usepackage{algorithm,algpseudocode}
\usepackage{amsmath}
\usepackage{amssymb} 
\usepackage{textcomp}
\usepackage{hyperref}
\usepackage[capitalize]{cleveref}
\usepackage{makecell}
\usepackage{booktabs}
\usepackage[dvipsnames]{xcolor}
\usepackage{tablefootnote}

\crefname{equation}{}{}

\newcommand{\specialcell}[2][c]{%
  \begin{tabular}[#1]{@{}c@{}}#2\end{tabular}}

\title{\LARGE \bf
Composing Dextrous Grasping and In-hand Manipulation via Scoring with a Reinforcement Learning Critic
}
\author{Lennart Röstel$^{*}$ \;\; Dominik Winkelbauer$^{*}$ \;\; Johannes Pitz \;\; Leon Sievers \;\; Berthold Bäuml%
\thanks{
    $^*$Equal contribution. The authors are with the Learning AI for Dextrous Robots Lab, Technical University of Munich, Germany (\href{https://aidx-lab.org/}{\scriptsize\texttt{aidx-lab.org}}), and the DLR Institute of Robotics \& Mechatronics (German Aerospace Center). 
    Contact: \{lennart.roestel, dominik.winkelbauer\}@tum.de\newline
}
}

\begin{document}

\maketitle
\pagestyle{empty}

\begin{abstract}
 In-hand manipulation and grasping are fundamental yet often separately addressed tasks in robotics.
 For deriving in-hand manipulation policies, reinforcement learning has recently shown great success. 
 However, the derived controllers are not yet useful in real-world scenarios because they often require a human operator to place the objects in suitable initial (grasping) states.
 Finding stable grasps that also promote the desired in-hand manipulation goal is an open problem.
 
 In this work, we propose a method for bridging this gap by leveraging the critic network of a reinforcement learning agent trained for in-hand manipulation to score and select initial grasps.
 Our experiments show that this method significantly increases the success rate of in-hand manipulation without requiring additional training.
 We also present an implementation of a full grasp manipulation pipeline on a real-world system, enabling autonomous grasping and reorientation even of unwieldy objects.
 Website: \href{https://aidx-lab.org/manipulation/icra25}{\scriptsize\texttt{aidx-lab.org/manipulation/icra25}}
\end{abstract}

\section{Introduction}
\label{sec:intro}

Dextrous grasping and in-hand manipulation with multi-fingered hands are challenging tasks involving multi-contact states and high degrees of freedom. 
While the state-of-the-art for both problems recently benefited from learning-based methods~\cite{Winkelbauer2022-uh,Sievers2022}, they are still largely addressed independently: 
Grasping deals with predicting static hand-object configurations that are robust to external disturbances by optimizing for a suitable grasp metric~\cite{Winkelbauer2022-uh}. 
Meanwhile, in-hand manipulation is about dynamically reorienting the object towards a desired object configuration~\cite{OpenAI2018-yi, chen2021system, Pitz2023-ra}.

While a stable grasp is often a prerequisite for successful in-hand manipulation, optimally combining grasping and in-hand manipulation for end-to-end performance remains an open problem. 
In particular, a key question is how to select grasps that best promote downstream in-hand manipulation tasks.
Assessing this suitability of a given grasp for subsequent manipulation (sometimes referred to as \textit{manipulability}) is traditionally challenging as it depends on a myriad of factors like the initial object pose, the object's shape and physical properties, the desired object configuration, and the hand's kinematic capabilities.

In this work, we propose a simple but effective solution to this problem that uses the critic network from a reinforcement learning agent after training to select grasps suitable for subsequent in-hand manipulation. 
To our knowledge, we are the first to address this problem with such a general solution, enabling increased end-to-end performance of a grasp-manipulation pipeline. 
Moreover, we think the method described in this paper applies to a larger class of problems in robotics, where optimization over a set of candidate initial states post-training is possible.
\begin{figure}
    \centering
    \includegraphics[width=0.4\textwidth]{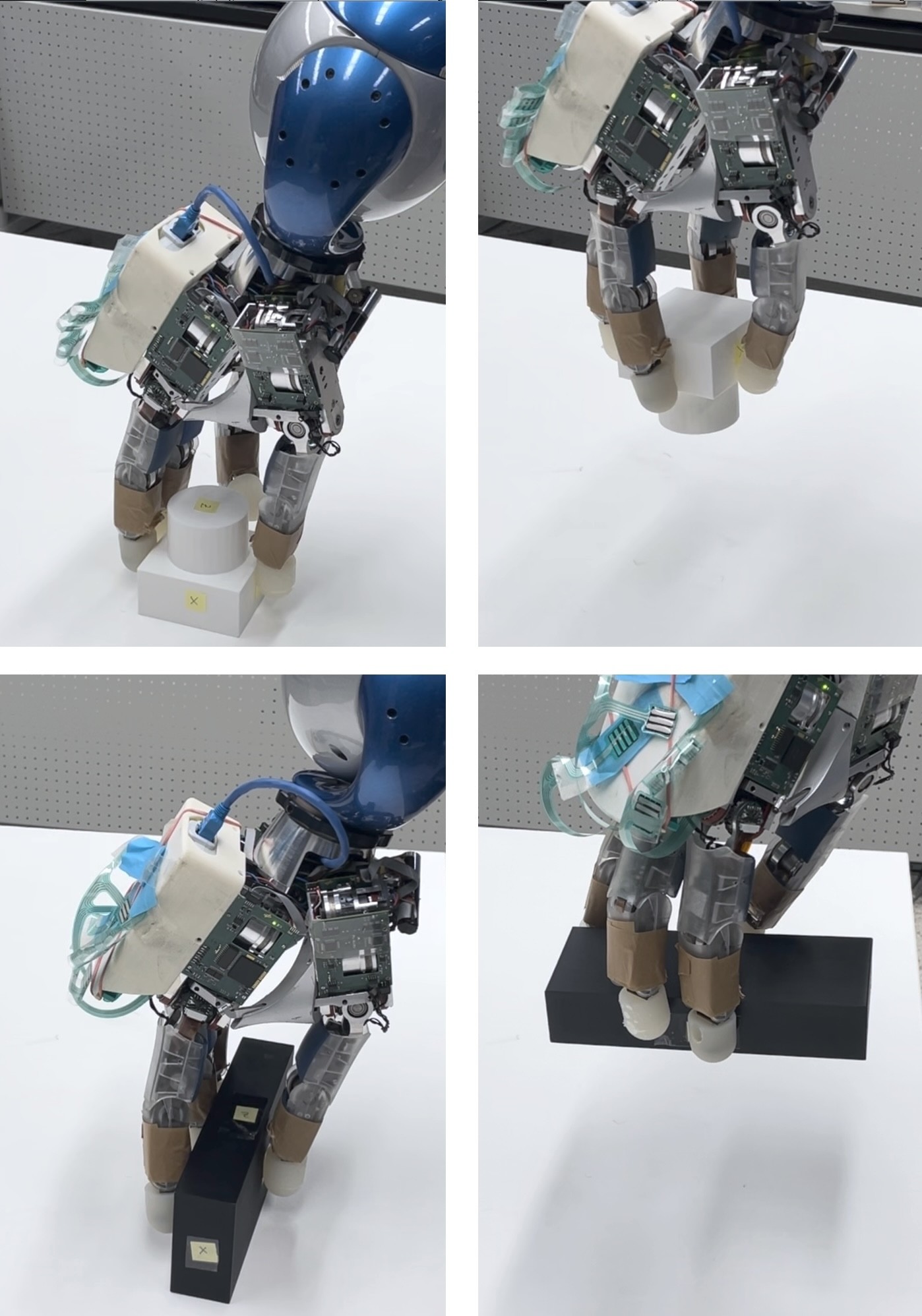}
    \caption{Grasping and in-hand manipulation of two objects with the DLR Hand II \cite{Butterfass2001}.
    Left: an object-specific grasp is selected, optimized to facilitate subsequent in-hand manipulation to the desired goal orientation (right).}
    \label{fig:title}
\end{figure}

\subsection{Related Work}
\label{sec:related_work}

\subsubsection{Grasping}

Grasping has been studied extensively in the past.
In early works, grasping is approached by analytical analysis using grasping properties like force closure \cite{Nguyen1988} or more fine-grained grasp qualities \cite{Li1988,Kirkpatrick1990, Ferrari1992}.
With the advent of data-driven approaches, grasps could be found efficiently online based on incomplete observations using SVMs \cite{Pelossof2004}, logistic regression \cite{Saxena2006}, or, more recently, deep-learning-based methods \cite{Lenz2015,Varley2015,Kappler2015}.
The latter allows high-dimensional input spaces without manual feature engineering and modeling complex grasp distributions for multi-fingered hands \cite{Li2022b,Wei2022,Winkelbauer2022-uh}.
Most mentioned methods are only concerned with the general grasp stability, while our approach further allows selecting grasps based on their suitability for the consecutive manipulation task.
Existing approaches that rate grasps based on their manipulability typically make use of the hand-object jacobian \cite{Roa2015}:
A manipulability measure can be defined via its condition number \cite{Salisbury1982}, minimum singular value \cite{Klein1987}, or the volume of the spanned manipulability ellipsoid \cite{Yoshikawa1985}.
While the listed methods were designed for serial manipulators, they can be transferred to multi-finger grasps \cite{Wen1999}.
Later, they were extended to support underactuated hands \cite{Prattichizzo2012} or to allow more fine-grained requirement specification \cite{Sato2011}.
While the mentioned approaches allow a more task-specific grasp selection, they only consider the grasp in a quasi-static setting.
In contrast, our approach rates grasps based on their suitability for the full subsequent manipulation task, including dynamic effects and finger gaiting.

\subsubsection{In-hand Manipulation}

\citet{OpenAI2018-yi} for the first time successfully applied deep reinforcement learning to reorient objects via finger gaiting. 
The seminal work showed that learning-based methods could overcome the challenges of modeling fragile multi-contact dynamics and long-horizon planning which classical approaches \cite{Okamura2000,Saut2007,Morgan2022} have struggled with.
Since then the field has seen impressive progress with simpler hardware setups, shorter training times \cite{Sievers2022,Qi2022-va,Khandate23,Rostel2023-nc}, for a specific object~\cite{openai2019rubiks,Handa2022-rc,Pitz2023-ra,Rostel2023-nc,Khandate24} or for manipulating multiple different objects, including objects unseen during training \cite{Chen2023-mq,Qi2023-th,Yuan23,wang2024,Pitz2024-zl}.
In all of these works, objects are placed in the hand by a human operator, limiting their applicability to real-world scenarios.
\citet{Chen2023-mq} additionally consider the task of manipulating objects that are placed on a table, however, without lifting them up.

In this work, we study in-hand manipulation initialized from diverse stable grasps that can be executed on a real robot system.
From an exploration perspective, a diverse distribution of initial states is known to be an important factor for successful reinforcement learning \cite{Khandate24}.
Additionally, having access to explicitly planned stable grasps (as opposed to learning an end-to-end policy for grasping and manipulation) allows to chain grasping, transportation, and in-hand manipulation in a controlled and generalizable manner.

Existing work on joint optimization of grasping and manipulation is limited to reduced settings like planar manipulators \cite{Horowitz2012}.
For the challenging case of multi-fingered hands, in this work for the first time we combine grasping and general in-hand reorientation in 3D in a principled way. 

\subsection{Contributions}
In summary, we present the following contributions:
\begin{itemize}
    \item We propose a novel method for scoring dextrous grasps, making use of a reinforcement learning critic network trained for in-hand manipulation.
    \item Utilizing grasp planning, as opposed to a heuristic across all objects, we can successfully train purely tactile in-hand $\mathrm{SO}(3)$-reorientation of objects with a larger aspect ratio than in any prior work. %
    \item We combine grasping and in-hand manipulation in a real-world pipeline on a humanoid robot, with generalization to objects unseen in training.
\end{itemize}

\section{Learning Dextrous Grasping and In-hand Manipulation}
\label{sec:components}
In this section, we describe the two main building blocks of our system: the candidate grasp generation \cref{sec:grasping} and the in-hand manipulation agent \cref{sec:manipulation}.
The former generates a set of candidate grasps for a given object, while the latter is trained to reorient objects in-hand using only tactile feedback, i.e., position and torque readings.
The methods described in this section are directly based on previous work on grasping \cite{Winkelbauer2022-uh, Humt2023-grasp} and purely tactile in-hand manipulation \cite{Rostel2023-nc, Sievers2022,Pitz2023-ra, Rostel2023-nc, Pitz2024-zl} and described in more detail in the respective papers.

\subsection{Candidate Grasp Generation}
\label{sec:grasping}

\begin{figure}[t!]
    \centering
    \vspace{3mm}
    \includegraphics[width=\linewidth]{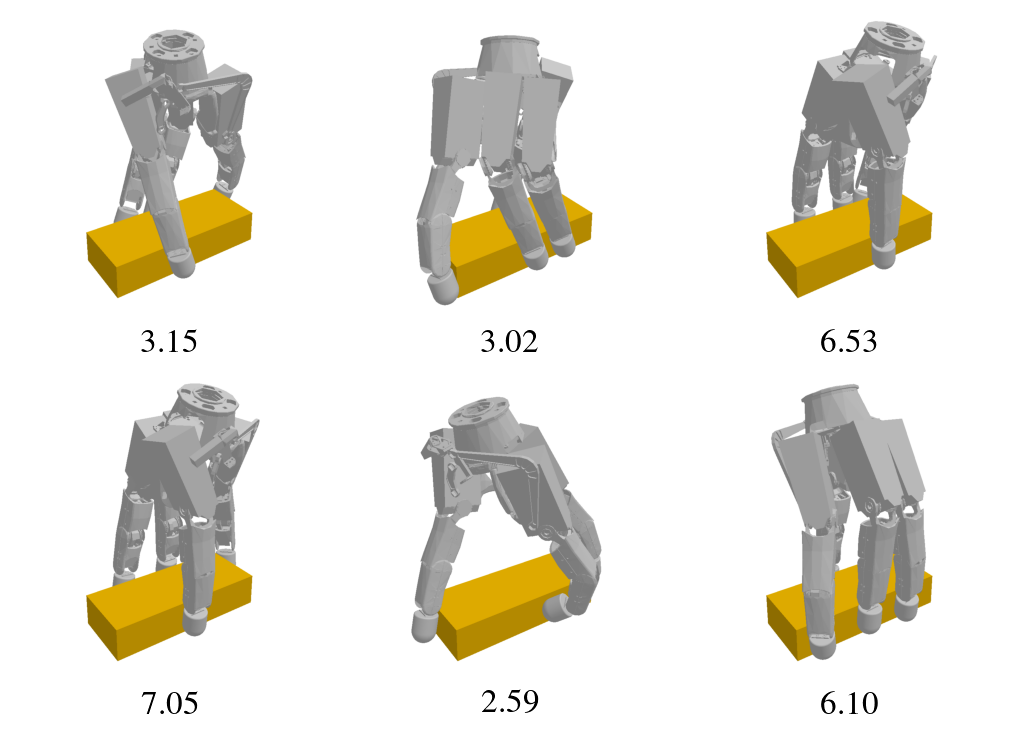}
    \caption{A collection of different candidate grasps on the same object. 
    The grasps shown were generated by our analytical grasp planner \cite{Winkelbauer2022-uh}. 
    The grasp quality (measuring robustness) is shown below each grasp.}
    \label{fig:grasps}
\end{figure}

The initial grasping step can be defined as finding a set of $K$ stable candidate grasps $\{[\mathcal{T}_H^{i}, q^{i}] \quad | i=1, ... K\}$ for a given object, with $\mathcal{T}_H^{i}$ being the hand pose and $q^{i}$ being the joint configuration of the $i$-th grasp.
Here, we make use of our learning-based grasping method described in \citet{Winkelbauer2022-uh} and \citet{Humt2023-grasp}.
This includes an analytical grasp planner which approaches grasping as an optimization problem with the objective being a grasp quality metric.
Specifically, we base our metric on the epsilon quality metric with additional adjustments to consider the hand kinematics, independent force application per finger, and uncertainty in the relative pose between hand and object.
Here, the quality metric can be interpreted as the minimum external wrench necessary to break the grasp, given that a fixed maximum torque budget is used on the finger joints.
This means the planned grasps are not task-specific, but instead are optimized to handle all possible external disturbances equally well.
See \cref{fig:grasps} for a random subset of grasps on the cuboid used throughout this work.
Using the analytical grasp planner, we generate a grasp dataset across different objects, which are then used to train our grasping network~\cite{Winkelbauer2022-uh}.
After training, the network can be used to efficiently generate arbitrary amounts of diverse grasps for a given object.
The input to the network consists of the shape completed object observed only via a single depth image, which allows the grasp generation stage to generalize to unknown objects.

\subsection{In-hand Manipulation Controller}
\label{sec:manipulation}
In line with prior work on tactile in-hand manipulation~\cite{Sievers2022,Pitz2023-ra,Rostel2023-nc, Pitz2024-zl}, we learn the in-hand reorientation controller by reinforcement learning in a realistic simulation environment. 
For training, we employ Estimator-Coupled Reinforcement Learning (ECRL)~\cite{Rostel2023-nc}, concurrently learning a policy $\pi$ and a state estimator $f$.
The state estimator $f$ recurrently estimates the object's pose $\hat{\mathcal{T}}_o^H$ from the history of joint measurements
\begin{equation}
    \label{eq:estimator}
 \hat{\mathcal{T}}_o^H(t + \delta t) = f\left(\hat{\mathcal{T}}_o^H(t), z(t), \mathcal{S}\right), 
\end{equation}
where $\mathcal{T}_o^H$ denotes the object pose in hand frame $H$.
The observation $z(t)$ is the concatenation of joint angle measurement $q$ and control targets $q_d$ stacked at a frequency of $60Hz$ for a window of $0.1$s.
The joint information enables the detection of contacts through an underlying high-fidelity impedance controller~\cite{Rostel2022-gu}. 
The shape encoding\footnote{The shape encoding vector is computed in each timestep as a function of the estimates object pose $\hat{\mathcal{T}}_o^H$ and object mesh $\mathcal{M}$. For readability, we suppress this explicit dependence and write $\mathcal{S}$ instead of $\mathcal{S}(\hat{\mathcal{T}}_o^H, \mathcal{M})$.}$S$ is obtained by transforming the canonical object mesh according to the estimated object pose $\hat{\mathcal{T}}_o^H$ and concatenating vectors between the object surface and a set of points fixed in $H$, as proposed in \citet{Pitz2024-zl}. 
We initialize the object pose at $t=0$ with a known pose obtained from a vision system (see \cref{sec:real}), subsequent estimates of object pose are done purely from tactile feedback.
The estimator $f$ is trained supervised to predict the object pose from sequences of joint measurements in simulation\cite{Rostel2023-nc}.

To achieve the in-hand reorientation, we train a goal-conditioned policy that outputs desired joint angles for the hand
\begin{equation}
    \label{eq:policy}
 q_d \sim \pi\left(\hat{R}_g^{o}, z, \mathcal{S}\right),
\end{equation}
where $\hat{R}_g^{o}$ is the rotation to the goal relative to the currently estimated orientation.
For learning the policy as well as the critic network, we use Proximal Policy Optimization (PPO)\cite{Schulman2017proximal}. 
The reward function is described in \cref{sec:scoring}.
Critically, during training, the hand object pose is initialized with a variety of grasps generated as described in \cref{sec:grasping}.
Similar to prior work \cite{Rostel2023-nc,Pitz2024-zl}, reorientations are considered successful if, after a time horizon of $\tau=10s$, the object orientation is within a threshold $\theta = 0.4$rad of the goal orientation.
When the goal is reached, a new goal orientation is sampled uniformly from $\mathrm{SO}(3)$, and the episode continues.
Training is conducted in a simulation environment based on the Isaac Sim physics engine~\cite{IsaacSim}. 

Suitable system identification and domain randomization of physical parameters (object masses, control gains, friction, contact parametrization, and measurement noise) enables a robust sim2real transfer as validated in prior work~\cite{Rostel2023-nc, Pitz2024-zl}.
We refer to \citet{Rostel2023-nc} for an in-depth description of the ECRL training procedure and to \citet{Pitz2024-zl} for the shape-conditioned policy and estimator architectures.

After initialization with a known object pose, the resulting agent is able to reorient objects without external supervision of the object pose, purely from tactile feedback (i.e., torque and position readings).
When autonomously deploying the agent, the policy is stopped when the estimated object orientation $\hat{R}_o^H$ is within a threshold of the goal orientation~$R_g^H$. 

\section{Composing Dextrous Grasping and In-hand Manipulation}
\label{sec:method}
This section outlines the core contribution of this paper: a novel method for maximizing the end-to-end performance of a grasp-manipulation pipeline. 
We first formulate an optimization problem over initial hand-object configurations, i.e., dextrous grasps.
We then show how the separately trained grasp generator and in-hand manipulation agent from \cref{sec:components} can be composed by scoring the generated grasps using the critic network trained for in-hand manipulation.

\subsection{Problem Formulation}
\label{sec:problem_formulation}
\begin{figure}[t!]
    \centering
    \vspace{3mm}
    \includegraphics[width=0.33\textwidth]{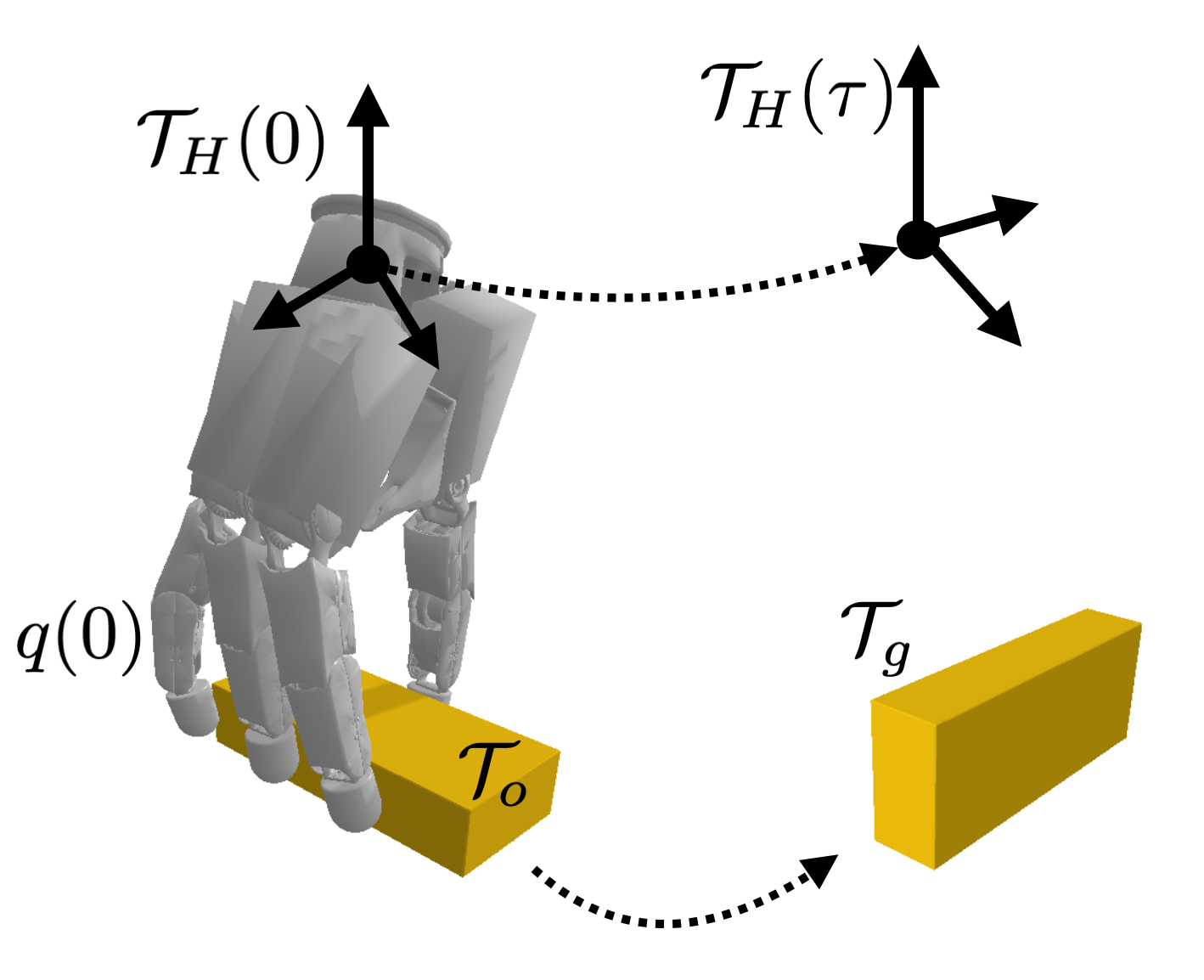}
    \caption{Illustration of the problem setup. 
    A multi-fingered hand is initially grasping an object (yellow) with pose $\mathcal{T}_o(0)$ using grasping joint angles $q(0)$ and is tasked with bringing it to a goal configuration $\mathcal{T}_g$ within time $\tau$.
    The hand pose $\mathcal{T}_H$ is controlled by a robotic arm (not shown). 
    When $\mathcal{T}_H$ is constrained due to the kinematics, the environment, or the task (e.g., the object needs to be placed on a table), arbitrary goal rotations of the object can only be achieved by in-hand manipulation.
    We are interested in finding $\mathcal{T}_H(0)$, $\mathcal{T}_H(\tau)$, and $q(0)$ that maximize the success probability of in-hand manipulation.}
    \label{fig:frame_vis}
\end{figure}
We study the problem of bringing an object with initial pose $\mathcal{T}_o(t=0) = \left[x_o(0), R_o(0)\right] \in \mathrm{SE}(3)$ to a goal configuration $\mathcal{T}_g = \left[x_g, R_g\right] \in \mathrm{SE}(3)$ in world frame using a multi-fingered hand mounted on a robotic arm (\cref{fig:frame_vis}). 
While the position component $x$ can be directly controlled by the robot arm to match $x_g$, the orientation component is typically not directly controllable and is most efficiently achieved by in-hand manipulation.
For this, we use an in-hand reorientation controller (see \cref{sec:manipulation}) that reorients an object from the initial object orientation $R^H_o(0)$ to $R_g^H$ in hand frame $H$ within time $t=\tau$ with success probability 
\begin{align}
    \label{def:success_rate}
 P\left(\text{success} \big| R_o^H(0), R_g^H, q(0), \mathcal{S}\right),
\end{align}

where ``success'' means that the angle between object orientation and goal orientation $d(R_o^H(\tau), R_g^H)$ is below a threshold $\theta$. 
In \cref{def:success_rate}, the probability of success additionally depends on the initial joint angles of the hand $q(0)$ and the object's shape $\mathcal{S}$.
The probabilistic formulation is motivated by the presence of inherent uncertainties in the system, e.g., due to sensor noise, unknown physical properties of the object and incomplete state information.

We are now interested in finding initial and final hand poses $\mathcal{T}_H(0)$ and $\mathcal{T}_H(\tau)$, as well as grasping joint angles $q(0)$ that maximize \cref{def:success_rate} subject to $\mathcal{T}_o(\tau) = \mathcal{T}_g$. 
Without loss of generality, for the experiments in this paper, we only consider hand poses $\mathcal{T}_H$ such that the hand is pointing downwards, which is amenable to grasping from and placing on flat surfaces.
Additionally, we only allow rotating the hand base by at most $\pi/2$ around the vertical axis to avoid twisted arm configurations and optionally impose a fixed wrist-rotation constraint $\mathcal{T}_H(0)=\mathcal{T}_H(\tau)$.

In this work, we require visual information only for inferring the initial state $\mathcal{T}_o(0)$ (and possibly the object's shape $\mathcal{S}$); for subsequent grasping and manipulation, we rely purely on tactile (i.e., torque-sensing) feedback of the hand.

\subsection{Optimizing Over Initial States by Critic-Scoring}
\begin{figure}[t!]
    \centering
    \vspace{3mm}
    \includegraphics[width=0.45\textwidth]{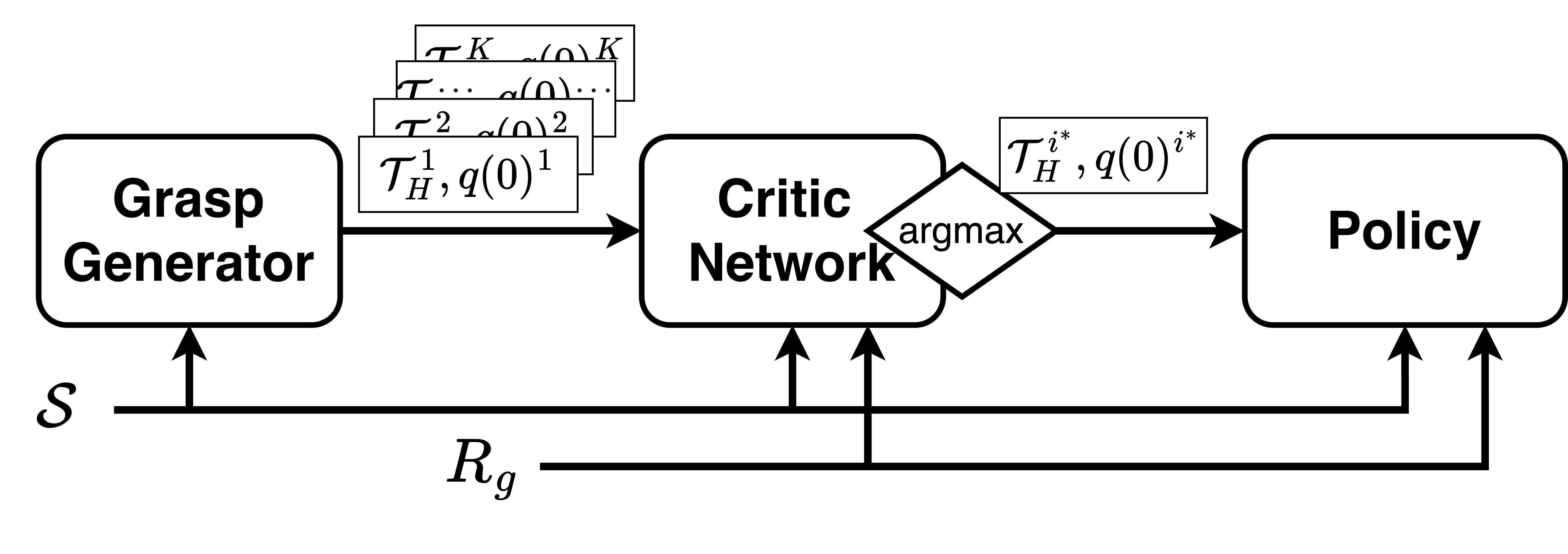}
    \caption{Scheme of the proposed method for selecting suitable grasps for in-hand manipulation. 
 First, a set of candidate grasps is generated for a given object shape $\mathcal{S}$ and pose $\mathcal{T}_o$.
 The grasp candidates $[\mathcal{T}_H^{i}, q^{i}]$ are generated w.r.t stability and satisfy applicable task constraints such as reachability and collision avoidance.
 Given a desired goal orientation $R_g$, the critic network is then used to score the grasps based on their suitability for achieving the desired rotation in-hand.
 The highest-scoring grasp is then executed and used to initialize the in-hand manipulation policy.
 }
    \label{fig:structure}
\end{figure}

In reinforcement learning, the value function $V(s_t)$ is defined as the expected discounted sum of rewards $r_t$ starting from state $s_t$ and following the policy $\pi$ thereafter
\begin{equation}
    \label{eq:value_function}
 V(s_t) = \mathbb{E}_{\pi}\left[\sum_{k = t} \gamma^{k-t} r_{k} \right]
\end{equation}
with discount factor $\gamma$.
In actor-critic methods like PPO, the value function is approximated by a neural network $v(s_t) \approx V(s_t)$ and learned concurrently with the policy by minimizing a temporal difference loss (see, e.g., \cite{Schulman2015-ua}).  
We note that for a sparse reward of the form
\begin{equation}
 r_{\text{s}}(t) = \begin{cases}
        1 & \text{if } t = \tau \quad \text{and} \quad \text{success} \\
        0 & \text{otherwise}, 
    \end{cases}
\end{equation}
a well-motivated assumption is that the value at $t=0$ 
\begin{equation}
    \label{eq:value_0}
 V(t=0) = \gamma^{\tau} \mathbb{E}_{\pi}\left[ r_{\text{s}}\left(t = \tau\right) \right],
\end{equation}
is a good proxy for the success probability of a given initial task configuration with normalization factor $\gamma^{\tau}$.

In the context of optimization over initial states $s(0)$, we, therefore, propose to use the learned critic network (as obtained from learning the in-hand manipulation policy in \cref{sec:manipulation}) for choosing the most suitable initial state for the task at hand.
A benefit of this post hoc optimization approach is that the initial states $s_i$ to be evaluated (referred to as \textit{candidate states}) can be chosen to match specific task constraints, which we will exploit in our application to grasp selection.
Moreover, evaluating batches of initial states is computationally efficient as it requires only a forward pass through the critic network.

\subsection{Scoring Grasps for In-Hand Manipulation}
\label{sec:scoring}
We now apply the critic-scoring approach described above to the problem of selecting suitable grasps for in-hand manipulation.
For learning in-hand manipulation, we use a reward composed of a dense and a sparse component $r = r_{\text{d}} + r_{\text{s}}$. 
The sparse reward is given after successful in-hand reorientation. 
Because training with only sparse rewards is challenging, we also use a dense reward that encourages rotating the object towards the goal while penalizing deviations in object position and joint position from their default values (see \citet{Pitz2024-zl}).
In practice, we parametrize the critic $v$ as a single Multi-Layer-Perceptron (MLP) with a two-dimensional output $(v_{\text{d}}, v_{\text{s}})$ estimating \cref{eq:value_function} for the dense and sparse reward terms separately.
While both outputs can be used for training the policy, only $v_{\text{s}}$ is used for scoring the initial states as a proxy for expected success \cref{eq:value_0}.
The input to the critic network is the same as for the policy \cref{eq:policy}.

Given a set of candidate grasps $\{[\mathcal{T}_H^{i}, q^{i}] \quad | i=1, ... K\}$ for initial object pose $\mathcal{T}_o(0)$ and a goal orientation $R_g$, we would like to find the most suitable grasps for subsequent in-hand manipulation, i.e., that maximizes \cref{def:success_rate}.
To this end, we score each grasp by evaluating the sparse component of the critic network $v_{\text{s}}$ for the initial state and choose the highest-scoring candidate 
\begin{equation}
    \label{eq:scoring}
 i^* = \arg\max_i \;v_{\text{s}}\left(\hat{R}_g^{o, i}(0), z^{i}(0), \mathcal{S} \right),
\end{equation}
where the object pose $\mathcal{T}_o^{H,i}$ and goal rotation $R_g^{H,i}$ are obtained by transformation of $\mathcal{T}_o$ and $R_g$ to the hand frame~$\mathcal{T}_H^{i}$.
Note that because the policy and critic expect inputs in hand frame $H$, the critic is agnostic to absolute hand-pose transformations.
This opens the possibility for separate optimization over initial and final hand poses $\mathcal{T}_H(0)$ and $\mathcal{T}_H(\tau)$ respectively (\textit{move base}), effectively changing only the goal orientation in hand frame $R_g^H$ for the same goal orientation in world frame $R_g$.

\section{Experiments}
\label{sec:results}
In this section, we evaluate the proposed method in simulation and on a real robot.
We assess the effect of grasp scoring on two in-hand hand manipulation policies: first, a shape-conditioned policy that is trained to reorient randomized geometric shapes, generalizing to new shapes at test time similar to \citet{Pitz2024-zl}.
Second, we train a policy to reorient a cuboid with dimensions $\SI{4}{cm} \times \SI{8}{cm} \times \SI{20}{cm}$, which represents to our knowledge the largest aspect ratio that has been considered for general $\mathrm{SO}(3)$ in-hand reorientation.

For a fair comparison to baselines, we only consider \textit{stable} initial grasps when assessing their suitability for in-hand reorientation. 

\subsection{Evaluation in Simulation}
\label{sec:sim_results}
In simulation, we conduct a series of evaluations to assess the effectiveness of the critic-scoring approach proposed in \cref{sec:method}.
For the experiments described in this section, we use a separate set of candidate grasps different from those used for training and leave domain randomization enabled during evaluation.

\subsubsection{\textbf{Correlation of Critic Score and Success Rate}}
\begin{figure}[t!]
    \centering
    \vspace{3mm}
    \includegraphics[width=\linewidth]{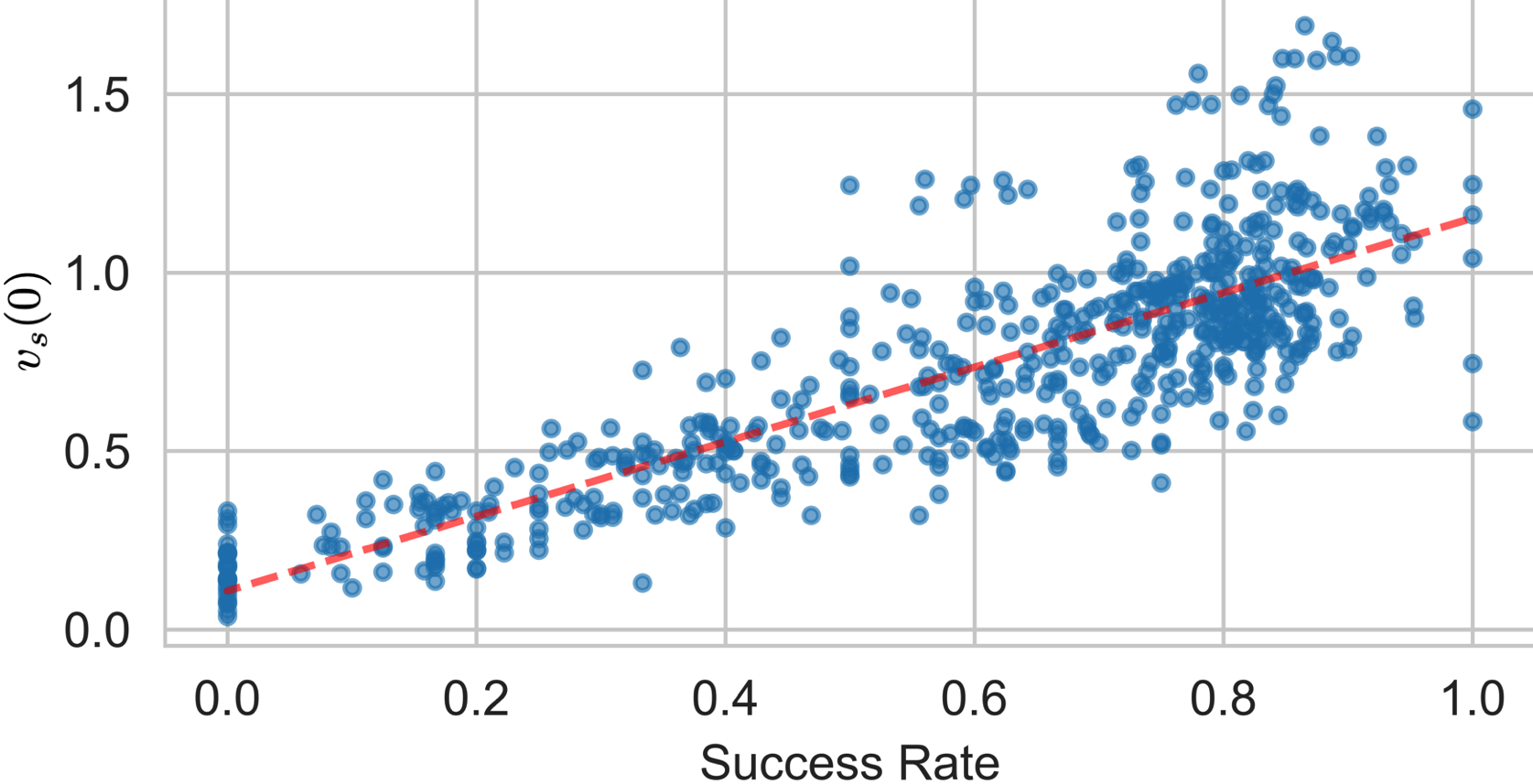}
    \caption{Correlation between critic score $v_{\text{s}}(t=0)$ and success rate of in-hand manipulation at timestep $\tau$ for the long cuboid object in simulation.
 Each data point corresponds to a pair of initial object rotation and goal rotation $(R_o(0), R_g)$, where $R_o(0)$ are aggregated within $0.3$rad to calculate the success rate.}
    \label{fig:correlation}
\end{figure}

We first show that the output of the critic network is predictive of in-hand manipulation success.
In \cref{fig:correlation}, we compare the value of $v_{\text{s}}(t=0)$ to the success rate of in-hand manipulation at timestep $\tau$ for various initial and goal rotations of the long cuboid object.

The data shows a clear correlation between the critic score and success rate with a Pearson correlation coefficient of~$0.82$.
We attribute the variance in the data mainly to the effect of noise injected into the simulation (domain randomization).

Note that the values of $v_{\text{s}}$ are not directly interpretable as probabilities here, even after normalization. 
This is expected, as the critic also takes into account the value of reaching goals after the horizon $\tau$ at multiples of $\tau$ thereafter (see \cref{sec:manipulation}).
Still, we expect, up to approximation errors and unobserved states, a monotonic relation between the success rate and the critic score \cref{eq:value_0}, justifying the use of the critic network for scoring grasps.

\subsubsection{\textbf{Effect of Selecting Grasps by Critic Score}}
\label{sec:effect_grasp_selection}
\begin{figure}[t!]
    \centering
    \vspace{3mm}
    \includegraphics[width=\linewidth]{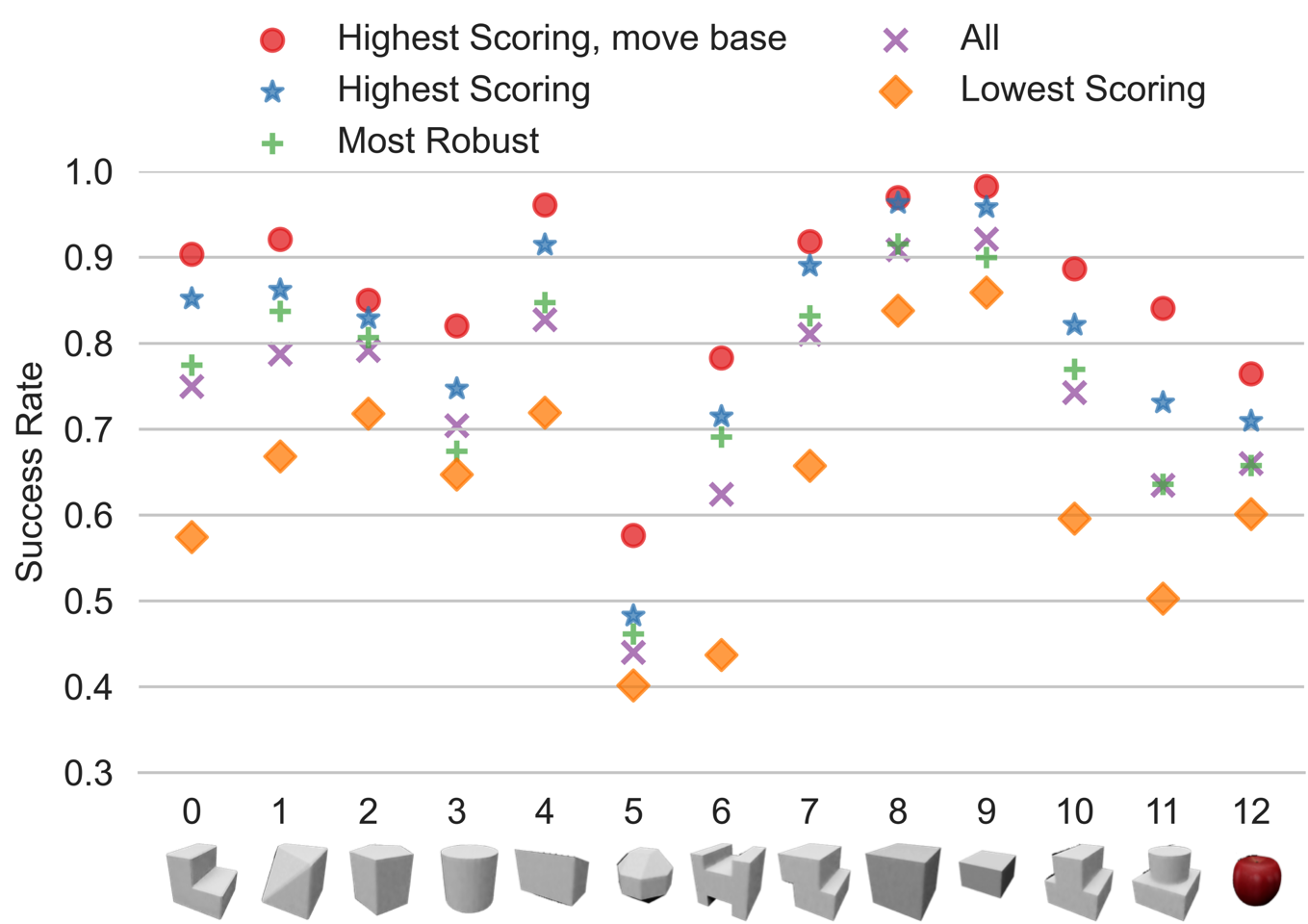}
    \caption{Success rates of in-hand manipulation in simulation for different strategies of choosing the initial grasp.
 Objects from \citet{Pitz2024-zl}, with objects 0-7 being in the training distribution and objects 8-12 out-of-distribution for the shape-conditioned policy.
 For each object, we simulate $N\approx50$k trials, leading to negligible statistical error.
 }
    \label{fig:baselines}
\end{figure}

We now assess the effect of grasp optimization on the in-hand manipulation performance.
To this end, we consider the following strategies for selecting the initial grasps among candidate grasps for a given object pose and goal orientation:
\begin{itemize}
    \item \textit{Most Robust}: Select the grasp with the highest epsilon quality metric.
    \item \textit{Highest Scoring}: Select the grasp with the highest critic score \cref{eq:scoring}, i.e., optimization over $q(0)$ and $\mathcal{T}_H(0) = \mathcal{T}_H(\tau)$.
    \item \textit{Highest Scoring, move base}: Select the grasp with the highest critic score while allowing the hand base to rotate around the vertical axis by at most $\pi/2$, i.e., in general $\mathcal{T}_H(0) \neq \mathcal{T}_H(\tau)$.
    \item \textit{Lowest Scoring}: Select the grasp with the lowest critic score (without base movement).
    \item \textit{All}: Average across all (stable) grasp candidates.
\end{itemize}
Note that the object rotation in world frame that needs to be accomplished by in-hand manipulation is unaffected by choice of grasp, except in the case of \textit{Highest Scoring, move base}.

In figure \cref{fig:baselines}, we show in-hand manipulation success rates for different strategies of selecting the initial grasp.
Across different objects, we find that selecting grasps based on the critic score consistently increases the success rate of in-hand manipulation compared to the average grasp or selecting grasps based on the epsilon quality metric.
Allowing the hand to rotate around the vertical axis as preferred by the critic network further increases the success rate.
\begin{table*}[t!]
    \caption {Effect of Grasp Selection Strategy on In-Hand Manipulation Performance in Simulation}
    \centering
    \begin{tabular}{l c c c}
        \toprule
        \textbf{Grasp Selection Strategy} & \specialcell{\textbf{Success Rate}\\ \textbf{Improvement [\%] $\uparrow$}} & \textbf{Object Dropped [\%] $\downarrow$} & \specialcell{\textbf{Average Time} \\ \textbf{to Goal [s]$\downarrow$}}\\
        \midrule
        All & - & 3.8 & 4.2 \\
        Most Robust & +2.3 & 2.9 & 4.2 \\
        \textbf{Highest-Scoring} & \textbf{+9.4} & \textbf{2.5} & \textbf{3.6}\\
        \textbf{Highest-Scoring, move base} & \textbf{+17.6} & \textbf{1.9} & \textbf{2.7} \\
        Lowest-Scoring & -14.6 & 7.8 & 4.7 \\
        \bottomrule
    \end{tabular}
    \label{tab:metrics}
\end{table*}
In \cref{tab:metrics}, we additionally report the improvements in success rate, the fraction of trials where the object is dropped during manipulation, and the time to reach the goal orientation. 
We aggregate all quantities over objects for the shape-conditioned policy with $N\approx600$k trials, leading to high statistical significance. 
While selecting the highest-scoring grasp already leads to a moderate decrease in both the fraction of objects dropped and runtime, the most prominent improvement is again observed when additionally allowing the hand base to rotate (\textit{Highest-Scoring, move base}). 
This can in-parts be explained by the ability of this strategy to partially compensate for in-hand reorientation by rotating the hand base (by at most $\pi/2$), which we find leads to an average decrease in required in-hand reorientation angle of approx. $\SI{0.4}{rad}$. 
For the large cuboid object, we observe a similar trend, with success rates of $78\%$ for the \textit{Most Robust}, $84\%$ for the \textit{Highest Scoring}, and $90\%$ for the \textit{Highest Scoring, move base} strategies.

\subsection{Real-World Validation}
\label{sec:real}
\begin{figure}[h]
    \centering
    \vspace{3mm}
    \includegraphics[width=\linewidth]{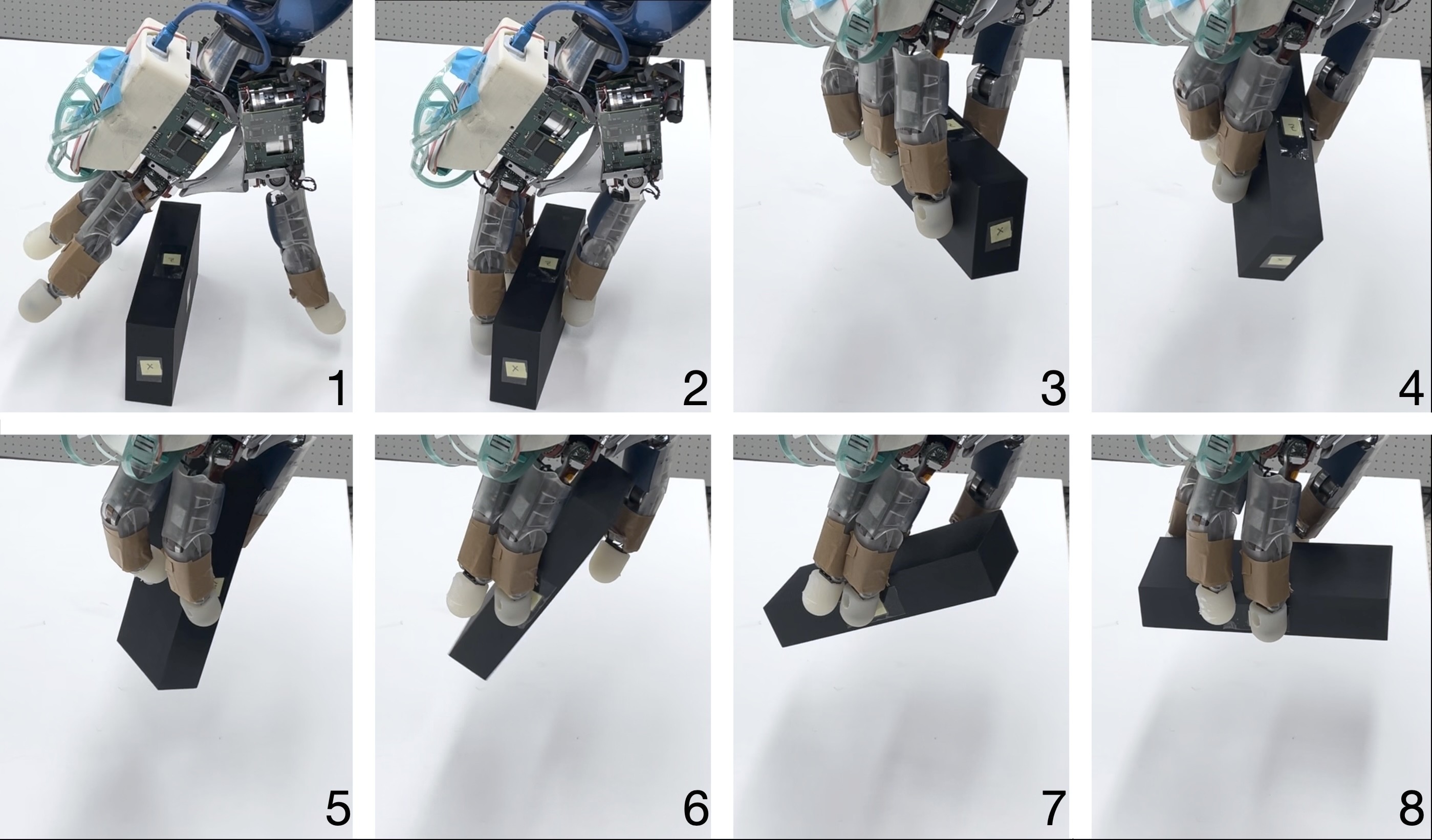}
    \caption{Real-world grasp-manipulation sequence for the large cuboid object.
    \textbf{1)} Pre-grasping \textbf{2)} Grasping pose with highest-scoring joint angles $q(0)$ and hand pose $\mathcal{T}_H(0)$ \textbf{3)} Arm lifts the object from the table and rotates to final hand pose $\mathcal{T}_H(\tau)$ (as predicted by critic scoring) without finger joint movement \textbf{4)}-\textbf{7)} In-hand manipulation by policy without supporting surface \textbf{8)} Goal orientation reached.
    }
    \label{fig:real_strip}
\end{figure}

The combination of grasping and in-hand manipulation is evaluated on our precisely calibrated~\cite{Tenhumberg2022b,Tenhumberg2023-Hand} humanoid robot \textit{Agile Justin}~\cite{Bauml2014-cr}.
The given object is observed as an incomplete point cloud using a Kinect depth camera~\cite{Wagner2013}.
Using our object-agnostic shape completion~\cite{Winkelbauer2022-uh}, a full 3D model is reconstructed and used to estimate the initial pose $\mathcal{T}_o(0)$ of the object.
This pose estimation is done via a global estimator using RANSAC registration based on FPFH feature matching~\cite{Rasu2009}, followed by a refinement stage based on the iterative closest point algorithm~\cite{Besl1992}.

Next, a set of \num{200} grasps $[\mathcal{T}_H^{i}(0), q^{i}]$ is predicted by our grasping network using the shape-completed 3D model and filtered based on collisions with 
the environment. Additionally, \num{90} reachable target hand poses $\mathcal{T}_H(\tau)$ are generated.
For each of the up to \num{18000} combinations, the critic network $v_s$ is evaluated, and the candidate combination $i^*$ of initial hand pose, final hand pose, and grasping joint configuration with maximum value is selected.
Evaluating the critic network for a batch of \num{18000} inputs takes $\sim$\SI{0.1}{s} on a T4 GPU, including the processing of the shape encoding.

Using a learning-based motion planner~\cite{Tenhumberg2022}, the hand now approaches the object as specified by the selected grasp $[\mathcal{T}_H^{i^*}, q^{i^*}]$.
Afterward, the fingers are closed until they contact the object, and the hand is lifted up into the final hand pose $\mathcal{T}_H^{i^*}(\tau)$. 

Now, the in-hand manipulation using the policy $\pi$ is started to reorient the object towards the given goal rotation $R_g^{H,i^*}$, without tracking of the object pose from the vision system (which would be impractical due to occlusions).
The policy stops autonomously when the object orientation $\hat{R}_o^H$ estimated from tactile feedback is within a threshold of the goal orientation $R_g^H$.
If the desired goal orientation was actually approximately reached, the trial is marked as a success.
In this procedure, two different objects are evaluated: the large cuboid, shown in \cref{fig:real_strip}, and an object that is out-of-distribution for the shape-conditioned policy, both shown in \cref{fig:title}.
For both objects, we exhaustively evaluate 24 goal rotations in a $\pi/2$-discretization of $\mathrm{SO}(3)$ (octahedral group), with 22 successful reorientations for the cuboid and 23 successes for the OOD object, representing an aggregated success rate of \num{93.7}\%.
Real-world rollouts for these and more objects can be found in the supplementary video.

\section{Conclusion}
\label{sec:conclusion}

In this work, we combine dexterous grasping and goal-conditioned in-hand object reorientation with a multi-fingered hand.
As the core idea, we propose the use of the critic network to choose grasps that maximize the probability of successful in-hand manipulation.
Our experiments showed that the proposed method substantially increases the success rate of in-hand manipulation compared to selecting grasps based on robustness alone.
We further demonstrated that the predicted values can be used to decide which movement of the robotic arm promotes the reorientation task.

A limitation of the current implementation is that the initial pose estimate obtained from vision can be ambiguous (e.g. due to occlusions).
In the future, we want to resolve such uncertainty by using tactile exploration during in-hand manipulation.
Another interesting direction for future work is to balance different objectives when optimizing grasps, including robustness and critic scores, to allow a more fine-grained, task-specific trade-off.
\bibliographystyle{IEEEtranN-modified}
\bibliography{IEEEabrv, bibliography}

\end{document}